\documentclass[10pt,twocolumn,letterpaper]{article}

\usepackage{cvpr}              
\definecolor{cvprblue}{rgb}{0.21,0.49,0.74}
\usepackage[pagebackref,breaklinks,colorlinks,allcolors=cvprblue]{hyperref}

\newcommand{\myparagraph}[1]{\smallskip\noindent\textbf{#1.}}

\usepackage{lineno}
\usepackage{multirow}

\usepackage{graphicx}
\usepackage{booktabs}
\usepackage{caption}
\usepackage[skins]{tcolorbox}
\usepackage[accsupp]{axessibility}  
\usepackage{tabularx}
\usepackage{soul}
\usepackage{enumitem}
\usepackage{newverbs}
\usepackage{subcaption} 
\usepackage{mathabx}
\usepackage{xcolor}
\usepackage{algorithmic}
\usepackage{algorithm}

\newtcbox\fp{hbox, on line, colback=LimeGreen, enhanced, frame hidden, boxrule=0pt, top=-2pt, bottom=-2pt, right=-2pt, left=-2pt, sharp corners}
    
\newtcbox\secp{hbox, on line, colback=Goldenrod, enhanced, frame hidden, boxrule=0pt, top=-2pt, bottom=-2pt, right=-2pt, left=-2pt, sharp corners}

\newtcbox\tp{hbox, on line, colback=SkyBlue, enhanced, frame hidden, boxrule=0pt, top=-2pt, bottom=-2pt, right=-2pt, left=-2pt, sharp corners}

\newtcbox\dsimp{hbox, on line, colback=Orchid, enhanced, frame hidden, boxrule=0pt, top=-2pt, bottom=-2pt, right=-2pt, left=-2pt, sharp corners}

\def\paperID{19476} 
\def\confName{CVPR}
\def\confYear{2026}

\title{
Scenes as Tokens: Multi-Scale Normal Distributions Transform Tokenizer for General 3D Vision–Language Understanding
}

\author{
Yutao Tang\footnotemark[1]
\textsuperscript{~1}, Cheng Zhao\textsuperscript{2}, Gaurav Mittal\textsuperscript{2}, Rohith Kukkala\textsuperscript{2}, \\ 
Rama Chellappa\textsuperscript{1},
Cheng Peng\textsuperscript{1}, Mei Chen\textsuperscript{2} \\
\textsuperscript{1}Johns Hopkins University, \textsuperscript{2}Microsoft \\
{\tt\small \{ytang67, rchella4, cpeng26\}@jhu.edu}, \\
{\tt\small \{cheng.zhao, gaurav.mittal, rohith.kukkala, mei.chen\}@microsoft.com}
}

\begin{document}
\twocolumn[
\maketitle

\begin{center}
\vspace{-1em}
\includegraphics[width=\textwidth]{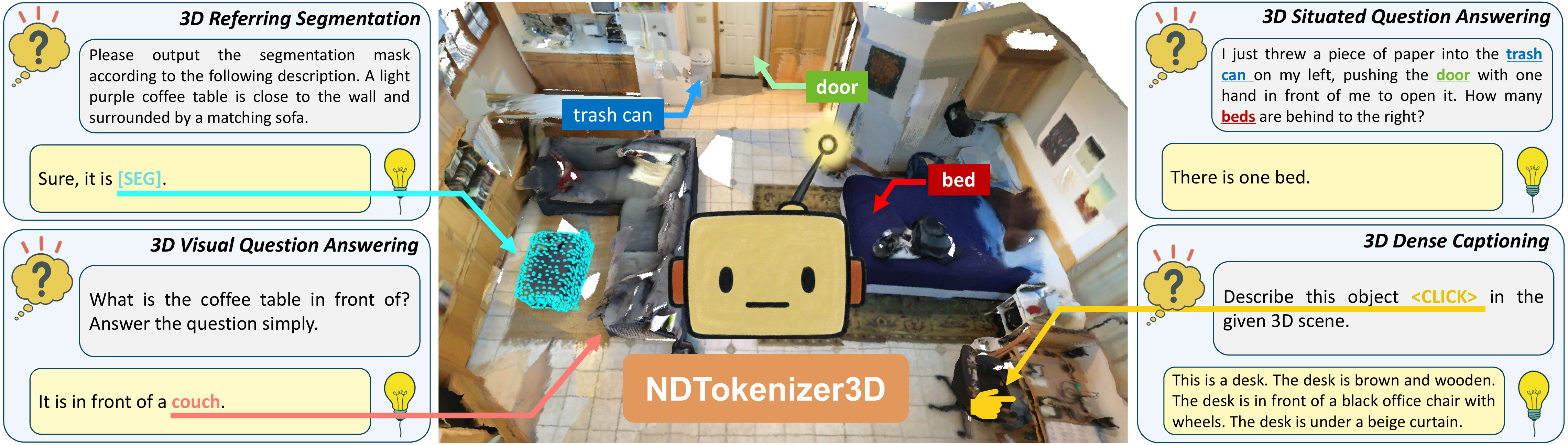}
\captionof{figure}{We introduce NDTokenizer3D, a generalist 3D VLM that bridges language-level reasoning with spatial understanding. By tokenizing complex 3D scenes into information-rich representations, NDTokenizer3D enables diverse tasks such as 3D Visual Question Answering, Dense captioning, and Referring Segmentation within a unified and interactive framework.
}
\label{fig:intro_figure}
\vspace{1em}
\end{center}

]

\footnotetext[1]{This work was done during Yutao's internship at Microsoft.}

\begin{abstract}
Recent advances in 3D vision–language models (VLMs) highlight a strong potential for 3D scene understanding and reasoning.
However, effectively tokenizing 3D scenes into holistic scene tokens, and leveraging these tokens across diverse 3D understanding tasks, remain highly challenging.
We present NDTokenizer3D, a generalist 3D VLM that performs a wide range of 3D scene understanding tasks while naturally supporting human interactions, thereby bridging language-level reasoning with 3D spatial understanding.
The core of our approach is a novel three-stage scene tokenization pipeline built upon a Multi-Scale Normal Distributions Transform (NDT) representation, paired with a Multi-Scale NDT Decoder (MSDec).
Specifically, NDTokenizer3D first constructs a multi-scale NDT representation from raw high-resolution point clouds, preserving both global context and fine-grained geometric details.
Next, the MSDec progressively fuses cross-scale NDT features, producing holistic scene tokens consumable by LLM endpoints.
Beyond tokenization, MSDec is repurposed as a general interface for human-interactive prompting (points, boxes, masks) and segmentation-mask decoding, unifying diverse 3D scene understanding tasks within a single architecture.
With this compact and unified design, NDTokenizer3D offers a fine-grained, general-purpose 3D VLM, achieving remarkable improvements in 3D Referring Segmentation, 3D Visual Question Answering, and 3D Dense Captioning.
\end{abstract}    
\vspace{-1.2em}
\section{Introduction}
\label{sec:intro}
Extending vision–language modeling from 2D to 3D opens the door to a range of impactful applications, such as autonomous driving, embodied AI , and 3D AR/VR, in which robust reasoning, fine-grained 3D scene understanding, and seamless human–agent interaction are essential.
Unlike conventional 2D VLMs, 3D VLMs require an effective 3D scene tokenization method to process 3D data. 
The core challenge lies in compressing massive high-resolution point clouds into bounded-length token sequences for an LLM endpoint while minimizing information loss.
Previously reported point-cloud–based tokenization pipelines~\cite{chen2024ll3da,zhi2025lscenellm,fu2025scenellm} process raw point clouds through downsampling, sacrificing fine-grained 3D geometric details and lacking an effective mechanism to capture the abstract global structure.
Moreover, achieving comprehensive 3D scene understanding requires modeling object-environment and inter-object relationships across multiple scales, as objects naturally appear at varying spatial resolutions. Consequently, it is crucial to capture both local details and broad contextual cues for holistic understanding. However, most current approaches either treat an entire 3D scene as a single-scale entity~\cite{deng20253dllava,chen2024ll3da} or represent a scene using a series of object instances~\cite{huang2024chatscene, chen2024grounded3dllm}, thereby neglecting the intricate multi-scale relationships that underpin realistic spatial reasoning.

To address these issues, we adopt a compact, memory-efficient multi-scale Normal Distributions Transform (NDT)~\cite{takeuchi2006ndt} representation that operates directly on high-resolution point clouds without downsampling. 
Originally developed for Simultaneous Localization and Mapping (SLAM), NDT partitions a 3D point cloud into a uniform grid and models each cell’s local surface as a Gaussian distribution. 
This grid-based formulation naturally supports multiple resolutions: fine-resolution cells capture detailed local geometry, while coarse-resolution cells encode global scene structure. 
Unlike naive downsampling, the fine-resolution NDT retains information from raw points via Gaussian statistics (mean and covariance) inside the NDT cell, simultaneously reducing memory usage and preserving local geometric details maximally. 
In parallel, the coarse-resolution NDT aggregates over larger spatial regions, providing an abstract representation that captures region-level global context information.
To effectively integrate these multi-scale representations, we further design a Multi-Scale NDT Decoder (MSDec) that progressively fuses features from coarse to fine scales, producing holistic scene tokens consumable by LLM endpoints.


Another challenge of 3D VLM lies in the limited support for human interaction~\cite{zhi2025lscenellm, huang2025reason3d} and the task-specific nature of existing methods~\cite{huang2025plm, he2024segpoint, saxena2025llmrg, man2024sig3d},
which fundamentally restricts their ability to generalize across diverse scene understanding tasks.
To overcome this limitation, we discard any task-tailored modules and instead deploy MSDec as a multi-purpose interface that supports both human interactive prompting and segmentation decoding within a streamlined framework. Concretely, MSDec processes user-provided visual prompts, such as points, bounding boxes, and masks, and embeds them into a guidance token, input to LLM along with the scene tokens for interactive 3D scene understanding. In addition, MSDec can also consume a special segmentation query token, from which it derives a 3D-aware segmentation feature that is decoded to produce segmentation masks.


Combining the strengths developed above, we present NDTokenizer3D, a generalist 3D VLM capable of performing a wide range of 3D scene understanding tasks while naturally supporting human interactions, bridging language-level reasoning with spatial understanding. NDTokenizer3D introduces a novel three-stage scene tokenization pipeline that represents 3D environments as information-preserving scene tokens. 
In stage one, NDTokenizer3D constructs a multi-scale NDT representation that captures both fine-grained local geometric details and abstract global contextual cues. 
In stage two, a 3D encoder extracts multi-scale features from these complementary NDT representations. 
In stage three, the extracted features are integrated through the MSDec, which progressively fuses information across scales to generate holistic scene tokens, enabling comprehensive reasoning over complex 3D environments. 
Beside tokenization, MSDec is repurposed as a general interface that enables human interactivity and segmentation mask decoding within the same framework. 
It can interpret user-provided visual prompts and incorporate them into the reasoning process for interactive 3D scene understanding. 
Additionally, the same decoder processes segmentation query to generate fine-grained 3D mask predictions, extending language reasoning naturally into spatial prediction.
Our contributions are summarized as follows:
\begin{itemize}
\item We propose a novel three-stage scene tokenization pipeline that leverages a multi-scale NDT representation to preserve both global context and fine-grained geometry, and integrates them through a MSDec to yield information-rich scene tokens.
\item We develop NDTokenizer3D as a generalist 3D VLM that supports human interactive prompting and segmentation decoding through a multi-purpose interface, unifying diverse 3D scene understanding tasks within a single architecture. 
\item We demonstrate that NDTokenizer3D achieves strong performance across four 3D vision–language benchmarks, showcasing its effectiveness and versatility across text generation and spatial reasoning tasks.
\end{itemize}

\section{Related Work}

\begin{figure*}[!htb]
    \includegraphics[width=\textwidth]{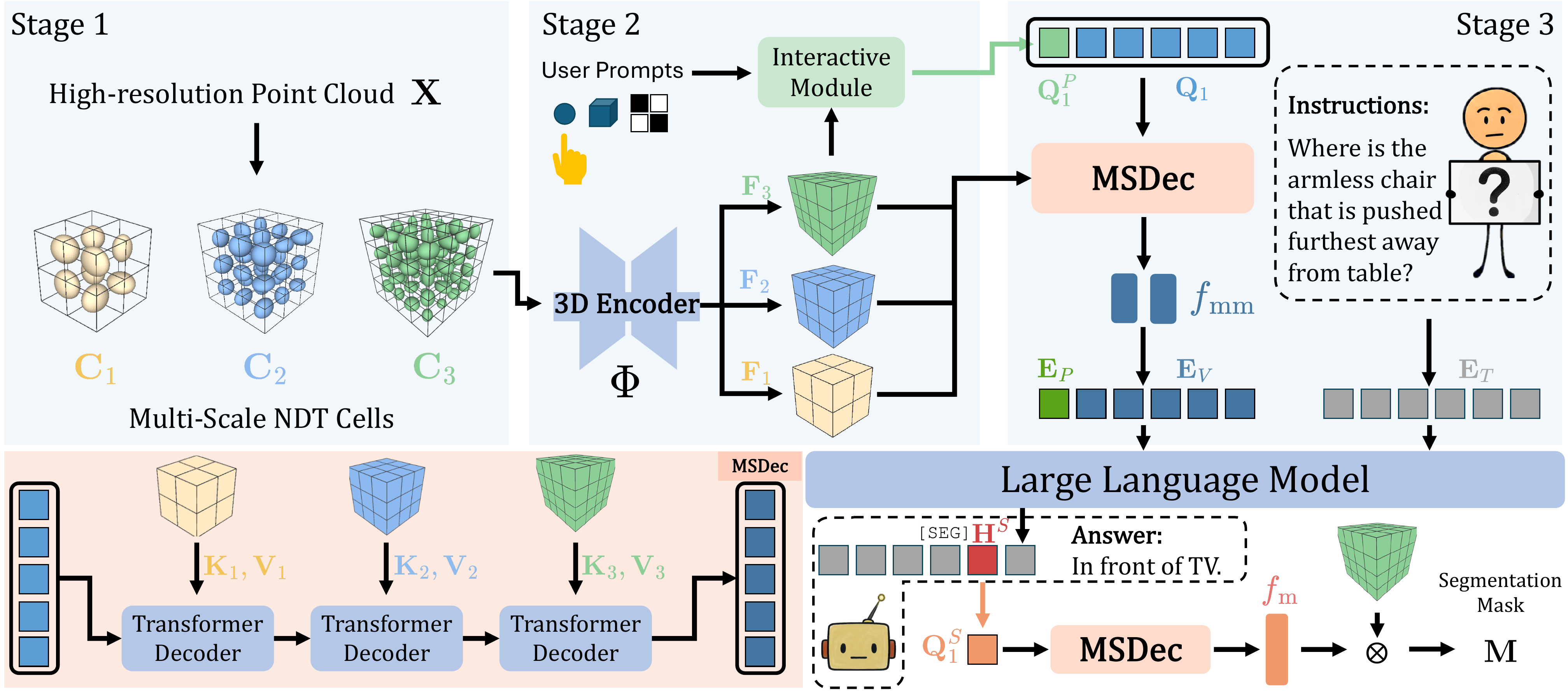}
    \caption{NDTokenizer3D is a general-purpose 3D VLM that supports a wide range of 3D understanding tasks.
    The model introduces a novel three-stage scene tokenization pipeline that constructs multi-scale NDT representations and aggregates them via MSDec to generate holistic scene tokens. The lower-left shows MSDec with $R$ transformer decoder layers that integrate multi-scale NDT features, using them as \texttt{Key} and \texttt{Value}. Beyond feature integration, MSDec also acts as a unified interface for user prompting and mask decoding.
    }
    \label{fig:pipeline}
    \vspace{-5mm}
\end{figure*}

\myparagraph{3D Scene Understanding}
Research at the intersection of vision and language has advanced a spectrum of 3D scene understanding tasks. Among them, 3D Visual Grounding~\cite{achlioptas2020nr3d,chen2020scanrefer,huang2022multi,yang2021sat,zhang2023multi3drefer} and Referring Segmentation~\cite{he2024segpoint,huang2025reason3d,qian2024x,wu20243d} focus on spatial understanding, where the goal is to localize or segment the object regions in 3D corresponding to natural language expressions. Moreover, 3D Question Answering~\cite{azuma2022scanqa,ma2022sqa3d,parelli2023clip} and Dense Captioning~\cite{chen2023end,chen2024vote2cap,chen2021scan2cap,jin2024tod3cap,kim2024bi,yuan2022x} emphasize semantic and relational reasoning. These tasks require the model to understand object attributes, spatial relations, and scene context to generate textual responses grounded in the 3D environment. 
%
Early progress in this field has largely focused on specialized architectures. Many approaches are single-task models~\cite{wu2023eda, chen2024vote2cap, jiao2022more, wang2022spatiality} designed for one specific objective, limiting their ability to generalize across diverse 3D understanding tasks. More recent efforts explore unified frameworks~\cite{cai20223djcg, chen2021d3net, chen2023unit3d} and pretraining strategies~\cite{jin2023context, zhu20233dvista, yang20243dvlp, zhang2024vision} that learn discriminative 3D features to enable shared reasoning across tasks.
Nevertheless, these systems typically follow a pre-train–then–fine-tune paradigm, relying on task-specific modules or per-task finetuning after pre-training. This reliance on extra components limits their flexibility and scalability, making it difficult for them to support more natural, human-like multimodal interaction and general-purpose 3D understanding.


\myparagraph{3D Vision Language Models}
Built upon 2D VLMs, many efforts extend them into the 3D domain by injecting geometric information into LLMs for 3D understanding.
A core challenge in this process lies in how 3D data are represented and aligned with language, so that the resulting features can be effectively interpreted and reasoned upon by the LLMs. Some approaches enhance 2D VLMs with limited 3D reasoning capabilities, \eg by introducing bird’s-eye view (BEV) image ~\cite{zhang2024agent3dzero,liu2025reasongrounder} or decomposing scenes into object sequences~\cite{ma2024spatialpin}. While convenient, such extensions lack full 3D awareness and lead to limited accuracy. Other works~\cite{fu2025scenellm, li20243dmit, huang2023chat3dv2} align 3D features to language through linear or MLP projections, whereas transformer-based modules~\cite{huang2023leo, yang2025lidarllm} provide richer modality interactions. More advanced methods~\cite{hong20233dllm, chen2024grounded3dllm} employ cross-modal query mechanisms such as Q-Former~\cite{li2023blip} to facilitate 3D–language fusion. Despite these advances, tokenizing 3D scenes into information-rich and LLM-consumable representations remains an open problem.
We address this challenge through a novel three-stage tokenization pipeline that produces globally informed, locally detailed 3D scene tokens for strong multimodal reasoning.

\section{Methods}

\Cref{fig:pipeline} presents an overview of the NDTokenizer3D framework, illustrated with a three-scale example.
It is designed as a general-purpose 3D VLM that flexibly integrates visual and textual inputs, enabling 3D vision-grounded dialogue, dense description, and referring segmentation. NDTokenizer3D begins by tokenizing 3D scenes into holistic scene tokens through a three-stage pipeline (\cref{sec:3d_scene_tokenization}). First, it constructs a multi-scale NDT representation from the high-resolution raw 3D point cloud as input (\cref{sec:msndt}). Next, a 3D Encoder extracts multi-scale 3D features from this representation (\cref{sec:feat_extraction}). Finally, a Multi-Scale NDT Decoder (MSDec) aggregates information across scales to produce holistic 3D scene tokens (\cref{sec:msdec}). In addition to feature aggregation, MSDec also serves as a unified interface that supports both user-input prompting and segmentation mask decoding (\cref{sec:msdec_unified_learner}).
Subsequently, we describe how the LLM utilizes all the tokens to produce task-specific outputs for various tasks in \cref{sec:llm_use_tokens}.
Finally, we present the training strategy in~\cref{sec:training_strategy}.

\subsection{3D Scene Tokenization}
\label{sec:3d_scene_tokenization}
\subsubsection{Multi-Scale NDT}
\label{sec:msndt}
Normal Distributions Transform \cite{takeuchi2006ndt} provides a compact and memory-efficient way to represent 3D scene data. It partitions a point cloud into regular cells and models the local surface within each cell as a Gaussian distribution. Leveraging this grid-based formulation, we construct a multi-scale NDT representation, where the fine-scale cells preserve detailed local geometric details, avoiding substantial information loss caused by aggressive downsampling, while the coarse-scale NDT cells captures abstract global semantics. Specifically, given the high-resolution point cloud $\mathbf{X} = \left\{  x_i\right\}_{i=1}^{N_p} \in \mathbb{R}^{N_p\times3}$, each NDT cell $C_r^j$ is characterized by its mean $\mu_r^j \in \mathbb{R}^{3}$ and covariance $\mathbf{\Sigma}_r^j \in \mathbb{R}^{3\times3}$:
\begin{equation}
\label{Eq:msndt}
\begin{gathered}
{\mu_r^j}= \frac{1}{n}\sum_{i=1}^n x_i, \ \mathbf{\Sigma}_r^j=\frac{1}{n-1}\sum_{i=1}^{n}(x_i - \mu_r^j)(x_i-\mu_r^j)^T,
\end{gathered}
\end{equation}
where $\left\{x_i \right\}_{i=1}^n $ are the 3D points inside each cell, and $r = 1, \cdots, R$ with $r=1$ representing the coarsest scale and $r=R$ the finest. These probability density functions implicitly interpret the structural and geometrical information of the local 3D points. 

\subsubsection{Feature Extraction}
\label{sec:feat_extraction}
In addition to the mean and covariance, we also incorporate RGB color information as part of each cell’s descriptor to enrich its visual representation. The cell-wise RGB values $c_r^j$ are obtained through multi-view projection from 2D images $\mathbf{I} = \{ I_k \}_{k=1}^{N_I}$ by
\begin{equation}
\label{eq:3d_2d_proj}
    c_r^j = \frac{1}{N_I} \sum_{k=1}^{N_I} I_k(u_k,v_k), \  [u_k,v_k]^T = P(\mu_r^j | k),
\end{equation}
where $k$ denotes a specific view, $P$ is the 3D-to-2D projective function, and $[u_k,v_k]^T$ are image coordinates. As a result, each cell is represented by the concatenation of its mean, covariance, and RGB values as
\begin{equation}
    \mathbf{C}_r = \left\{ C_r^j \right\}_{j=1}^{N_r} \in \mathbb{R}^{N_r \times 15}, \  C_r^j = \left[ \mu_r^j; \mathbf{\Sigma}_r^j; c_r^j \right] .
\end{equation}
After obtaining NDT cells, we employ a transformer-based 3D Encoder $\Phi$ to derive multi-scale NDT features $\mathbf{F}_r = \Phi(\mathbf{C}_r) \in \mathbb{R}^{N_r \times d_f}$ where $d_f$ is the dimension of the features. 
The coarse-scale features emphasize global context such as overall scene layout, environmental context, and inter-object relationships, while the fine-scale features encode detailed geometric and textural cues that describe intra-object structures and fine-grained boundaries. These encoded features serve as the basis for our multi-scale decoding framework, which integrates and refines information across different resolutions.

\subsubsection{Multi-Scale NDT Decoder}
\label{sec:msdec}
To effectively integrate information across different scales, we design a transformer-based Multi-Scale NDT Decoder (MSDec), which learns a holistic and semantically rich scene representation capturing both global context and local details essential for accurate 3D scene reasoning. Specifically, the MSDec consists of $R$ transformer decoder layers where we employ the multi-scale NDT features to serve as \verb|Key| and \verb|Value| inputs in each decoder layer, while a set of learnable queries serves as initial \verb|Query|. In the first decoder layer, we initialize the query tokens using a downsampled subset of the finest-scale features by 
$\mathbf{Q}_1 = \mathbf{W}^Q_1 (\downarrow {\mathbf{F}}_R)$
where $\mathbf{W}^Q_1$ is a linear projection and $\downarrow$ denotes the downsample operator. Each MSDec layer then updates the queries through a cross-attention layer with multi-scale NDT features, followed by a self-attention layer and a feed-forward network (FFN) layer as
\begin{equation}
\begin{gathered}
    \tilde{\mathbf{Q}}_r = \operatorname{CrossAttn}(\mathbf{Q}_r,\mathbf{K}_r,\mathbf{V}_r)\\
    \hat{\mathbf{Q}}_{r} = \operatorname{SelfAttn}(\tilde{\mathbf{Q}}_r), ~~
    \mathbf{Q}_{r+1} = 
    \operatorname{FFN}(\hat{\mathbf{Q}}_{r}),
\end{gathered}
\end{equation}
where $\mathbf{K}_r = \mathbf{W}^K_r \mathbf{F}_r, \mathbf{V}_r = \mathbf{W}^V_r \mathbf{F}_r$. 
The final output $\mathbf{Q}_{R}$ encapsulates the holistic 3D scene representation, integrating information from all scales. This representation is then projected via a multimodal alignment head $f_{\text{mm}}$ into the LLM input space, yielding the 3D scene tokens $\mathbf{E}_V = f_{\text{mm}} (\mathbf{Q}_{R})$, serving as visual context. 
%
Through this hierarchical decoding process, MSDec effectively fuses global semantics with fine-grained details, tokenizing a 3D scene into information-rich tokens that enable context-aware 3D reasoning within the language model.

\subsection{MSDec as a Unified Interface}
\label{sec:msdec_unified_learner}

Beyond multi-scale feature integration, MSDec serves as a unified interface that supports both human prompting, enabling interactive 3D scene understanding, and segmentation mask decoding, which extends text generation to spatially grounded 3D prediction. The following sections detail how MSDec operates in each of these modes.

\myparagraph{User-Input Prompting}
A robust 3D VLM should support user-input prompts such as clicks, boxes, and masks. NDTokenizer3D handles these via MSDec, aligning user intent with 3D context for interactive understanding. Concretely, inside the interactive module, we first convert user input into a binary mask $m_u \in \mathbb{R}^{N_R \times 1}$ over the finest-scale features $\mathbf{F}_R$, and the masked region is average-pooled to yield the prompt feature $\mathbf{F}_R^P$. This feature initializes an additional query $\mathbf{Q}_1^P$, concatenated with the original query set $\mathbf{Q}_1$ and jointly processed through all MSDec layers. The resulting prompt-conditioned output $\mathbf{Q}_R^P$ is then projected through the same $f_{\text{mm}}$ to form a guidance token $\mathbf{E}_P$. By incorporating user-input prompts directly at the decoder level, MSDec effectively aligns user intentions with the multi-scale 3D context, enabling flexible, interaction-driven 3D understanding while maintaining spatial and semantic coherence.

\myparagraph{Segmentation Mask Decoding}
Beyond user prompting, MSDec also supports segmentation mask decoding as part of its unified design. When instructed to perform a segmentation task, the LLM generates a special token \verb|[SEG]|. The hidden state of the \verb|[SEG]| token $\mathbf{H}^S$ is projected through a segmentation head $f_{\text{s}}$ to form a query $\mathbf{Q}_1^{S}$, which MSDec decodes into a segmentation-aware representation $\mathbf{Q}_R^{S}$ using the same decoding procedure as in the previous section. A mask head $f_{\text{m}}$ then transforms $\mathbf{Q}_R^{S}$ into a kernel that interacts with the finest-scale features via a dot product to predict the final 3D mask $\mathbf{M} \in \mathbb{R}^{N_R \times 1}$.
This design allows segmentation to emerge naturally from the language reasoning process, rather than being treated as a separate prediction branch or a post-processing step, thereby maintaining architectural coherence and modality alignment within the unified framework.

\subsection{Multimodal Token Integration}
\label{sec:llm_use_tokens}
After our scene tokenization pipeline and the multi-purpose decoding interface, our system produces a set of visual tokens that serve as the visual context for the language model. 
The final outputs of MSDec include two types of visual tokens: (1) the scene tokens $\mathbf{E}_V$ representing holistic 3D scene information, and (2) the guidance token $\mathbf{E}_P$ derived from user-interactive cues when available. The textual instructions are tokenized using the pretrained LLM tokenizer to obtain $\mathbf{E}_T$. All tokens are concatenated to form the multimodal input sequence for the LLM, yielding
\begin{equation}
    \hat{a} = \operatorname{LLM} ([\mathbf{E}_V; \mathbf{E}_P; \mathbf{E}_T]), \ 
    \texttt{[SEG]} \subset \hat{a},
\end{equation}
where $\hat{a}$ denotes the generated textual response. The LLM can generate the special token \verb|[SEG]| within the response; when present, it triggers the mask decoding process described in \cref{sec:msdec_unified_learner}.

\subsection{Training Strategy}
\label{sec:training_strategy}

\myparagraph{Stage 1: Pre-Training 3D Encoder and MSDec} Since there are no pre-trained weights designed for NDT representations, we first pre-train the 3D Encoder and the MSDec jointly on the 3D Instance Segmentation task. To this end, we append two heads after MSDec: (1) a multi-class classification head for predicting the semantic category of each instance, and (2) a mask head $f_{\text{m}}$, identical to that described in \cref{sec:msdec_unified_learner}, for generating instance masks. We use a combination of cross-entropy loss $\mathcal{L}_{cls}$ for classification and a segmentation loss $\mathcal{L}_{m}$ for mask prediction, where $\mathcal{L}_{m}$ consists of binary cross-entropy and Dice loss \cite{milletari2016diceloss}. To further encourage the encoder to produce semantically aligned 3D features, we incorporate 2D visual-language supervision. Specifically, we extract 2D image features using CLIP \cite{radford2021clip} image encoder and lift them into the 3D space following the projection procedure described in \cref{eq:3d_2d_proj}, obtaining $\mathbf{F}_r^{\text{C}} \in \mathbb{R}^{N_r \times d_f}$. Then, we apply a cosine similarity loss between $\mathbf{F}_r^{\text{C}}$ and $\mathbf{F}_r$ as
\begin{equation}
\label{eq:cos_sim}
    \mathcal{L}_s(\mathbf{F}_r^{\text{C}}, \mathbf{F}_r) = \frac{1}{N_r} \sum_{j=1}^{N_r} 1 - \frac{\mathbf{F}_r^{j,\text{C}} \mathbf{F}_r^j}{||\mathbf{F}_r^{j,\text{C}}||_2 ||\mathbf{F}^j_r||_2}.
\end{equation}
The final loss for Stage 1 is then given by 
\begin{equation}
    \mathcal{L} = \mathcal{L}_{cls} + \lambda_1 \mathcal{L}_m + \lambda_2 \mathcal{L}_s(\mathbf{F}_r^{\text{C}}, \mathbf{F}_r).
\end{equation}

\myparagraph{Stage 2: Instruction Tuning} In this stage, we freeze 3D Encoder and MSDec while training only the projection layers, \ie $f_{\text{mm}}$ and $f_{\text{s}}$, along with the LLM. We conduct multi-task training using a combined dataset comprising Referring Segmentation, Visual Question Answering, and Dense Captioning tasks. Besides the next-token generation cross-entropy loss $\mathcal{L}_{{t}}$, we include the same mask prediction loss $\mathcal{L}_{{m}}$ as described in Stage 1. Furthermore, to encourage the LLM to produce semantically consistent responses, we introduce a cosine similarity loss between the hidden states $\mathbf{H}^{\hat{a}}$ of the predicted answer $\hat{a}$ and the text embeddings $\mathbf{H}^a$ of the ground-truth answer $a$ extracted by the CLIP \cite{radford2021clip} text encoder, as defined in \cref{eq:cos_sim}. The overall objective is then given by
\begin{equation}
    \mathcal{L} = \mathcal{L}_{t} + \lambda_3 \mathcal{L}_m + \lambda_4 \mathcal{L}_s(\mathbf{H}^{\hat{a}}, \mathbf{H}^a).
\end{equation}

\section{Experiments}
\label{sec:exp}

\myparagraph{Datasets}
We conduct experiments on the ScanNet dataset \cite{dai2017scannet}, including 1201 training scenes and 312 validation scenes. In Stage 1, we utilize the instance mask annotations from ScanNet200 \cite{rozenberszki2022scannet200}, an extended version of ScanNet that introduces more fine-grained semantic categories for 3D Instance Segmentation pre-training. In Stage 2, we perform multi-task instruction tuning using a combined dataset of approximately 295k instruction–response pairs. Specifically, we train on ScanRefer \cite{chen2020scanrefer}, Nr3D \cite{achlioptas2020nr3d}, and Multi3DRefer \cite{zhang2023multi3drefer} for Referring Segmentation, where the model localizes target objects based on textual descriptions, and evaluate on the Multi3DRefer validation set. For Visual Question Answering, which requires reasoning over 3D scenes to answer natural language questions, 
we train on ScanQA \cite{azuma2022scanqa} and SQA3D \cite{ma2022sqa3d}, and evaluate on their respective validation and test sets. Lastly, for Dense Captioning, which aims to generate region- and object-level descriptions within a 3D scene, we train on Nr3D \cite{achlioptas2020nr3d} and Scan2Cap \cite{chen2021scan2cap}, and evaluate on the Scan2Cap validation set.
Additionally, we evaluate hallucination performance on the 3D-POPE benchmark \cite{yang20253dgrand}, which measures a model’s ability to correctly judge the presence of objects within a 3D scene. 

\begin{table*}[t]
    \setlength{\belowcaptionskip}{-0.6em}
    
    \begin{minipage}[c]{0.63\textwidth}
        \centering
        \setlength{\tabcolsep}{1.4pt}
        \resizebox{\textwidth}{!}{
        \begin{tabular}{l |c| c c c c| c c| c c c c}
            \hline
            {\multirow{2}{*}{Generalist 3D VLMs}}  & Multi3DRefer & \multicolumn{4}{c|}{ScanQA} & \multicolumn{2}{c|}{SQA3D} & \multicolumn{4}{c}{Scan2Cap} \\
            \cline{2-2} \cline{3-6} \cline{7-8} \cline{9-12}
            &  mIoU & C & B-4 & M & R & EM & EM-R & C@0.5 & B-4@0.5 & M@0.5 & R@0.5 \\
            
            \hline

            Spatial 3D-LLM \cite{wang2025spatial3dllm} &-	&82.5	&13.9	&16.8	&39.1	&46.2	&-	&72.2	&34.6	&23.1	&54.3 \\
            LEO \cite{huang2023leo}  & - & {\color{gray}101.4} & {\color{gray}13.2} & {\color{gray}20.0} & {\color{gray}49.2} & 50.0 & 52.4 & 72.4 & \textbf{38.2} & \underline{27.9} & \textbf{58.1} \\
            Scene-LLM \cite{fu2025scenellm}  & - & 80.0 & 11.7 & 15.8 & 35.9 & 53.6 & - & - & - & - & - \\
            Chat-Scene \cite{huang2024chatscene}  & - & 87.7 & 14.3 & 18.0 & 41.6 & \textbf{54.6} & \textbf{57.5} & 77.2 & 36.4 & \textbf{28.0} & \textbf{58.1} \\
            Grounded 3D-LLM \cite{chen2024grounded3dllm}  & - & 72.7 & 13.4 & - & - & - & - & 70.6 & 35.5 & - & - \\
            Oryx MLLM \cite{liu2024oryx} & - & 74.3 & - & 15.3 & 38.4 & - & - & - & - & - & - \\
            LSceneLLM \cite{zhi2025lscenellm} & - & 88.2 & - & 18.0 & 40.8 & - & - & - & - & - & - \\

            3D-LLaVA \cite{deng20253dllava}   & 42.7 & \underline{92.6} & \textbf{17.1} & \underline{18.4} & \underline{43.1} & \underline{54.5} & 56.6 & \underline{78.8} & \underline{36.9} & 27.1 & \underline{57.7} \\


            NDTokenizer3D (Ours)  & \textbf{46.0}	&\textbf{98.6}	& \underline{17.0}	&\textbf{19.4}	&{\textbf{44.9}}	&54.4&	\underline{57.1}	&\textbf{79.0}	& 36.7	&27.1&	\underline{57.7} \\
            
            \hline
            
        \end{tabular}
        }
    \end{minipage}
    \hfill 
    \begin{minipage}[c]{0.37\textwidth}
        \centering
        \includegraphics[width=\linewidth]{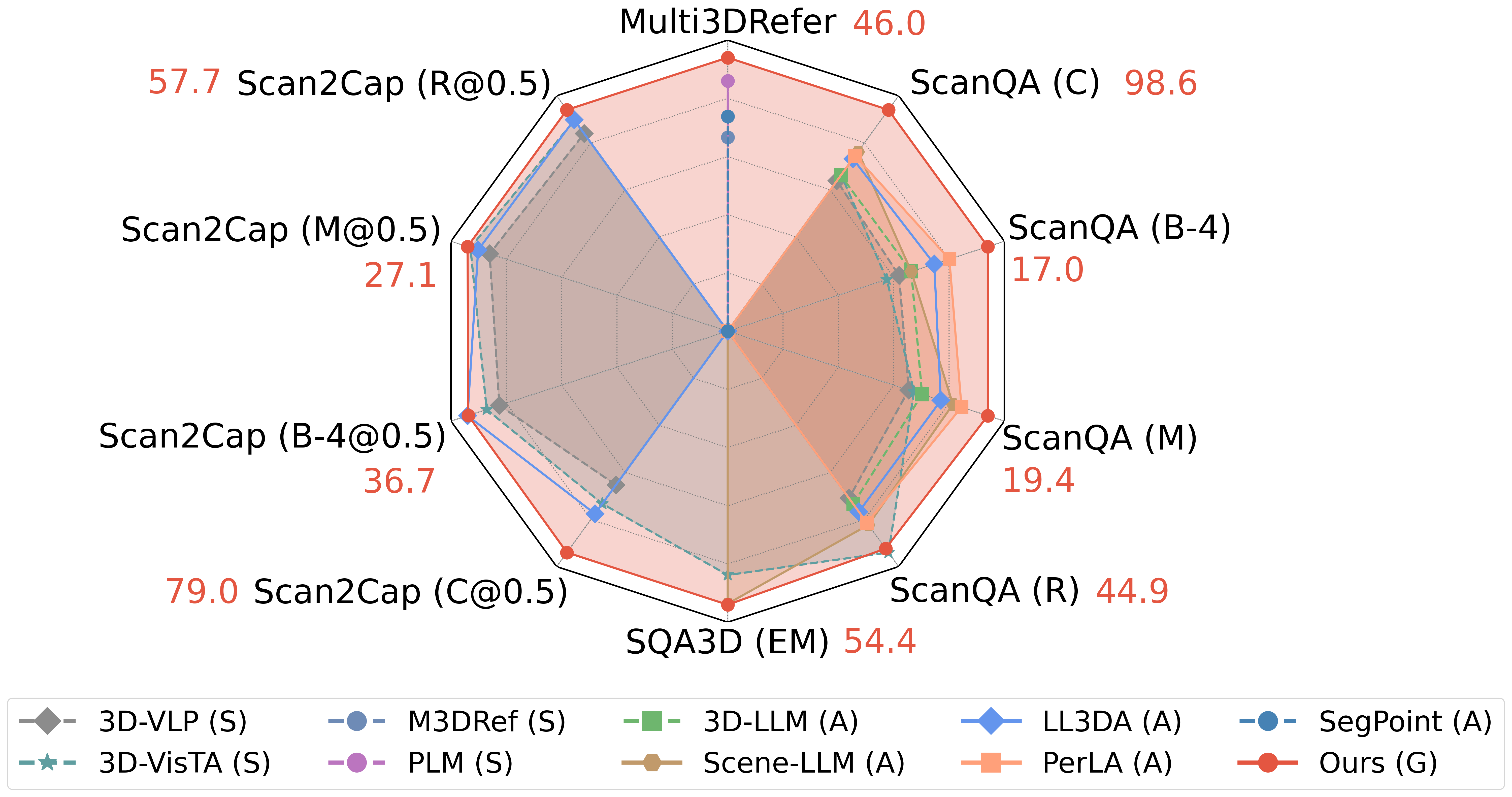}
    \end{minipage}
    

\caption{\textbf{Left:} Quantitative comparison of state-of-the-art generalist 3D VLMs. Results for LEO \cite{huang2023leo} on ScanQA are shown in gray as it operates under a different setting with access to ground-truth objects and are thus not directly comparable. The best and second-best results for each metric are \textbf{bold} and \underline{underlined}, respectively. 
NDTokenizer achieves the strongest overall performance, particularly on segmentation and QA tasks.
\textbf{Right:} Radar plot comparing the per-task performance of specialist models and adapted 3D VLMs. NDTokenizer consistently surpasses both categories across all tasks.
}
\label{tab:main_results_radar_plot}

\vspace{-3mm}

\end{table*}

\myparagraph{Metrics}
We use mean intersection over union (mIoU) as the evaluation metric~\cite{deng20253dllava} for Referring Segmentation, and report extract match accuracy (EM) and its refined version (EM-R) for SQA3D. Following prior works \cite{deng20253dllava, zhi2025lscenellm, huang2024chatscene}, we adopt CiDEr (C) \cite{vedantam2015cider}, BLEU-4 (B-4) \cite{papineni2002bleu}, METEOR (M) \cite{banerjee2005meteor}, and Rouge-L (R) \cite{lin2004rouge} to evaluate the quality of generated text responses on ScanQA and Scan2Cap.

\myparagraph{Implementation Details}
We use Point Transformer v3 (PTv3) \cite{wu2024ptv3} as the 3D Encoder and employ FlashAttention-2 \cite{dao2023flashattention} to enhance memory and computational efficiency within MSDec. MSDec takes 850 initial learnable queries. The multimodal alignment head, the segmentation head, and the multi-class classification head are implemented as two-layer MLPs. For the LLM endpoint, we adopt the language model from the LLaVA-1.5-7B model \cite{liu2024improved} in our framework. Training is conducted on 4× NVIDIA A100 GPUs with the DeepSpeed toolkit \cite{rasley2020deepspeed} for distributed optimization. We apply LoRA fine-tuning \cite{hu2022lora} on the LLM.
We perform instruction tuning for one epoch. The batch size is set to 32 per GPU. We use the AdamW optimizer \cite{loshchilov2017adamw} with a cosine annealing learning rate schedule, starting from an initial learning rate of $8 \times 10^{-5}$ and a warm-up ratio of 10\%. 
Additional implementation details are provided in the Supplementary Materials.

\subsection{Quantitative Comparisons}
We compare NDTokenizer3D with existing approaches in \cref{tab:main_results_radar_plot}. The models compared are categorized into three groups: specialist models (S), adapted 3D VLMs (A), and generalist 3D VLMs (G). Specialist models are designed for specific tasks and lack generalization, while adapted 3D VLMs require task-specific fine-tuning that improves in-domain performance but limits cross-task transfer. 
In contrast, generalist 3D VLMs are trained on unified multi-task datasets, enabling a single model to perform diverse 3D vision–language tasks. The left panel of \cref{tab:main_results_radar_plot} summarizes the comparison among generalist 3D VLMs, while the right shows results for specialist and adapted models. Comparison with more methods and full metrics is provided in the Supplementary Materials.

\myparagraph{3D Referring Segmentation}
Multi3DRefer \cite{zhang2023multi3drefer} benchmark measures a model’s ability to localize a flexible number of objects in real-world 3D scenes from natural language descriptions by predicting 3D masks. Each description in Multi3DRefer may refer to zero, one, or multiple target objects. 
As shown in \cref{tab:main_results_radar_plot}, only our NDTokenizer3D and 3D-LLaVA \cite{deng20253dllava} are capable of performing both language-centric tasks and point-level understanding tasks (\eg Referring Segmentation). Our NDTokenizer3D achieves 46.0 mIoU, surpassing 3D-LLaVA by \textbf{+3.3}. This improvement stems from two key design choices: (1) our multi-scale NDT representation preserves both local geometric details and global contextual information, whereas 3D-LLaVA relies on superpoint-pooling and token selection, effectively a form of downsampling; and (2) our MSDec integrates features across multiple scales, unlike 3D-LLaVA’s single-scale design. Together, these components enable more spatially coherent scene representations and more precise language-grounded segmentation.

\myparagraph{3D Visual Question Answering}
ScanQA \cite{azuma2022scanqa} evaluates a model’s 3D perception and reasoning capabilities across diverse question types, including object counting, identification, attribute recognition, and appearance reasoning. In contrast, SQA3D \cite{ma2022sqa3d} focuses on situated reasoning, requiring the model to infer spatial context, such as position, orientation, and viewpoint from text and reason about surrounding 3D environments.
Our NDTokenizer3D consistently achieves top performance on both datasets. Notably, on ScanQA, it attains 98.6 CiDEr and 19.4 METEOR, surpassing the second-best results by \textbf{+6.0} and \textbf{+1.0}, respectively. 
These results validate the strong reasoning and alignment capabilities of our proposed 3D scene tokenization pipeline when integrated with LLMs.

\myparagraph{3D Dense Captioning}
The Scan2Cap benchmark \cite{chen2021scan2cap} evaluates a model’s ability to describe a user-specified object’s appearance and its spatial relations with surrounding objects. Following prior works \cite{deng20253dllava, zhu20233dvista, huang2023leo}, we adopt object proposals generated by Mask3D \cite{schult2022mask3d} for evaluation, while notably not using Mask3D outputs during training. Our NDTokenizer3D treats these object proposals as visual prompts, which are processed by MSDec (\cref{sec:msdec_unified_learner}) to provide guided visual context to the LLM.
Our NDTokenizer3D achieves competitive results across all metrics, demonstrating strong capability in generating spatially coherent and contextually rich descriptions grounded in 3D environments. Moreover, this experiment highlights our model’s capability to interpret and incorporate high-level human intent directly into its reasoning process. This property is essential for achieving interactive 3D understanding, indicating that NDTokenizer3D goes beyond passive scene perception to support intent-driven multimodal reasoning.

\begin{table}[!htb]
    \setlength{\belowcaptionskip}{-1em}
    \centering
    \resizebox{0.47\textwidth}{!}{%
    \begin{tabular}{l|cc|cc|cc}
        \hline

         {\multirow{2}{*}{Methods}} &  \multicolumn{2}{c|}{Random} & \multicolumn{2}{c|}{Popular} & \multicolumn{2}{c}{Adversarial} \\
     \cline{2-3} \cline{4-5} \cline{6-7}
     & Prec & Acc & Prec & Acc & Prec & Acc   \\

\hline        
        
        LEO \cite{huang2023leo} & 51.95&	52.91&	48.30&	47.27&	48.47&	47.52 \\
        
        
        3D-LLaVA \cite{deng20253dllava} & \underline{73.24}	& \underline{80.32}	&\underline{66.86}	&\underline{74.11}	&\underline{63.13}	&\underline{69.88} \\

         Ours & \textbf{80.34}&	\textbf{84.12}&	\textbf{69.69}&	\textbf{75.51}&	\textbf{66.15}&	\textbf{72.03} \\

\hline
    \end{tabular}}

    \caption{Hallucination performance on 3D-POPE. Our NDTokenizer3D achieves the lowest hallucination rates across all settings.}
    \label{tab:3dpope}

\end{table}

\myparagraph{Hallucination Performance}
3D-POPE \cite{yang20253dgrand}
consists of 
yes/no questions about whether a specific object exists in the scene. This benchmark is balanced, containing an equal number of positive and negative questions.
To construct negative questions (nonexistent objects), 3D-POPE provides three different settings: (1) Random Sampling: randomly selects an object category that is absent in the current scene. This setting evaluates whether the model tends to hallucinate arbitrary objects without contextual grounding. (2) Popular Sampling: selects the most frequent object categories across the dataset that are not present in the current scene. This tests whether the model hallucinates common objects that it has seen often during training. (3) Adversarial Sampling: for each object that truly exists in the scene, it identifies objects that often co-occur with it in the training data but are absent in the current scene. The most strongly co-occurring absent object is then used as an adversarial negative sample. This design targets realistic contextual hallucinations rather than random or popular object confusions.
As shown in \cref{tab:3dpope}, our model exhibits substantially lower hallucination rates across all three settings, consistently outperforming previous generalist methods by a large margin. These results show that NDTokenizer3D provides more informative scene tokens that are faithful to the underlying geometry, leading to more accurately grounded LLM predictions.


\begin{figure*}[!htb]
    \setlength{\belowcaptionskip}{-1.2em}
    \setlength{\abovecaptionskip}{-0.2em}
    \includegraphics[width=\textwidth]{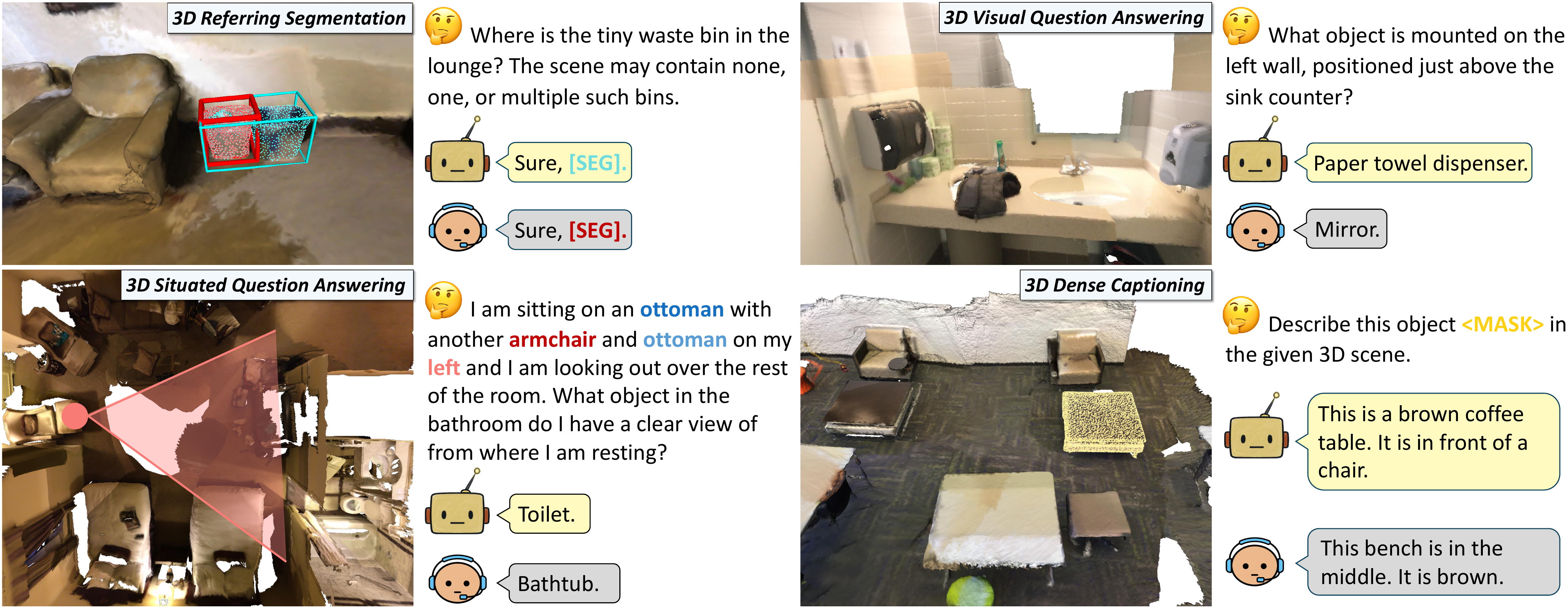}
    \caption{Qualitative comparison between NDTokenizer3D \includegraphics[height=1.5em]{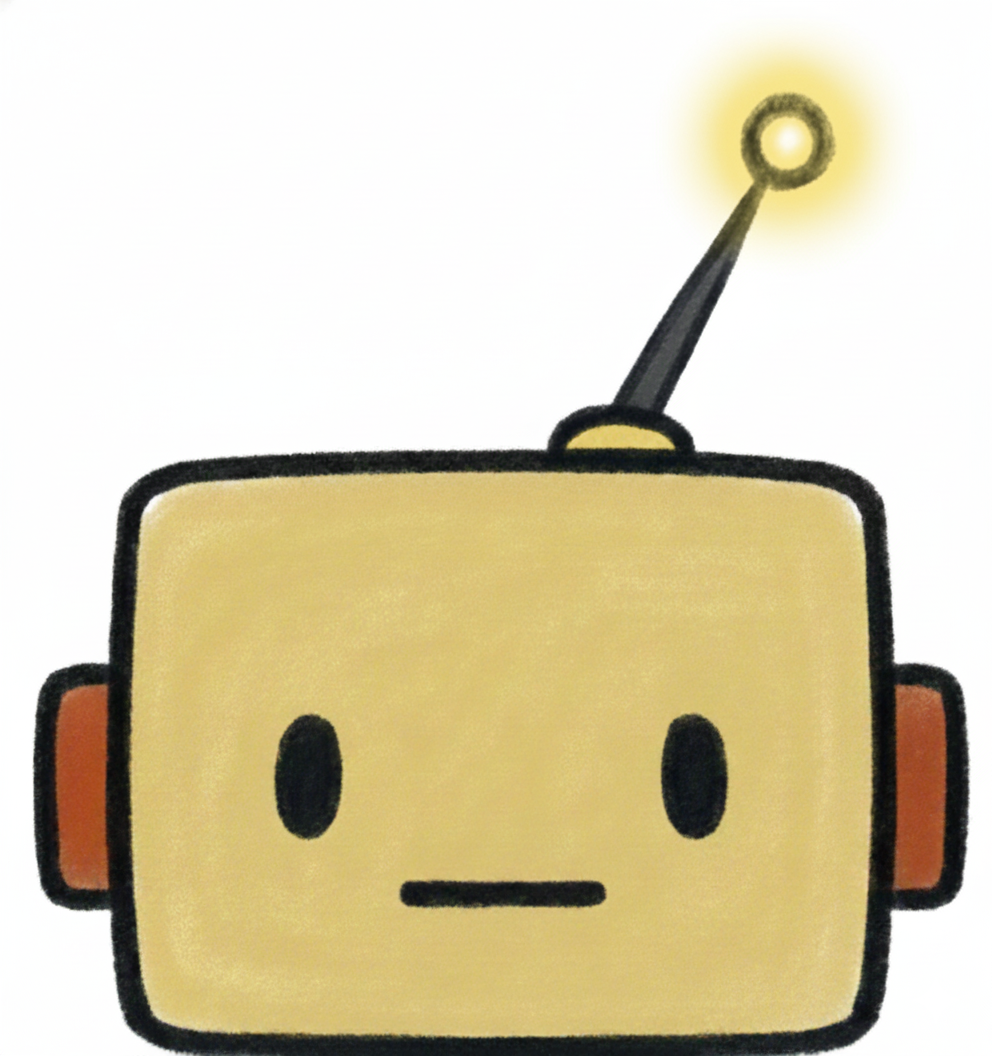} and 3D-LLaVA \includegraphics[height=1em]{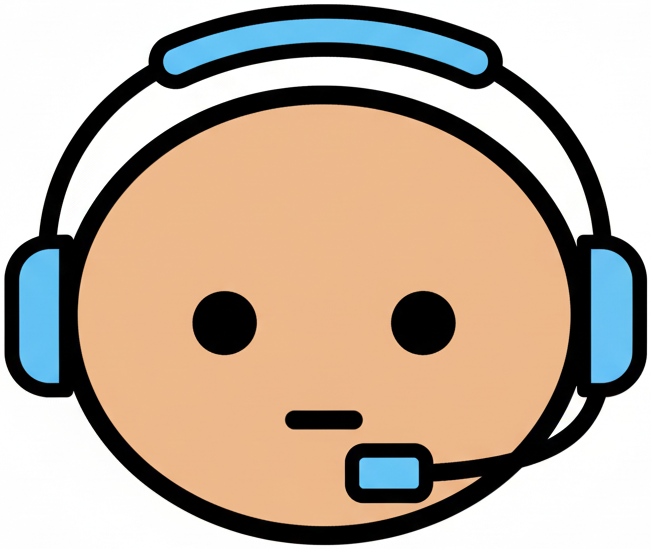} across four tasks, showing NDTokenizer3D's improved grounding, spatial reasoning, and object understanding.
    }
    \label{fig:qualitative}
\end{figure*}

\subsection{Ablation Studies}

\begin{table}
\centering
    \centering
    \setlength{\tabcolsep}{1.5pt}
    \resizebox{0.47\textwidth}{!}{

    \begin{tabular}{l | c |c c c c| c c| c }

     \hline
                 
     {\multirow{2}{*}{Methods}} & Multi3DRefer & \multicolumn{4}{c|}{ScanQA} & \multicolumn{2}{c|}{SQA3D} & {Scan2Cap} \\
     \cline{2-2} \cline{3-6} \cline{7-8} \cline{9-9}
     
     & mIoU & C & B-4 & M & R & EM & EM-R & C@0.5 \\

     \hline
    
    Baseline &  45.3	&94.7&	16.2&	18.8&	43.9&	53.9&	56.5&	78.1  \\
    Ours & \textbf{46.0}	&\textbf{98.6}&	\textbf{17.0}&	\textbf{19.4}&	\textbf{44.9}&	\textbf{54.4}&	\textbf{57.1}&	\textbf{79.0} \\

     \hline

    \end{tabular}}

    \caption{Ablation comparing multi-scale NDT compression with a downsampling baseline. Best socres are in \textbf{bold}. Our NDT-based tokenization consistently improves performance across all tasks.}
    \label{tab:ablation_downsampled_pcd}

\end{table}

\myparagraph{Comparison with Downsampling}
We conduct an ablation study comparing our proposed multi-scale NDT compression with a naive multi-scale downsampling baseline. Specifically, we replace the NDT cells with voxel-downsampled point clouds containing the same number of points as NDT cells at each scale, while keeping all other architectural components identical. Formally,
\begin{equation}
    \mathbf{P}_r = \left\{ P_r^j \right\}_{j=1}^{N_r} \in \mathbb{R}^{{N_r} \times 6}, \ 
    P_r^j = \left[ p_r^j; c_r^j \right],
\end{equation}
where $p_r^j$ and $c_r^j$ denote the 3D coordinates and RGB color, respectively.
The resulting multi-scale point sets $\{ \mathbf{P}_r \}_{r=1}^R$ are processed by the same tokenization pipeline and fed into the LLM for reasoning. Results are summarized in \cref{tab:ablation_downsampled_pcd}. Our method consistently outperforms the downsampled baseline across all tasks, confirming that NDT provides richer and more discriminative scene representations. On ScanQA in particular, replacing NDT with downsampled points leads to a noticeable drop, indicating that naive downsampling removes fine-grained geometric cues essential for correct reasoning. Interestingly, the baseline still performs competitively on Referring Segmentation, surpassing 3D-LLaVA \cite{deng20253dllava} (\cref{tab:main_results_radar_plot}), highlighting that multi-scale integration is highly beneficial for point-level understanding. A similar trend appears on ScanQA, where the baseline also outperforms 3D-LLaVA, showing that the multi-scale integration helps capture inter-object relations and broader contextual cues, even without NDT statistics.

\begin{table}
\centering
    \setlength{\belowcaptionskip}{-1em}
    \centering
    \setlength{\tabcolsep}{1.4pt}
    \resizebox{0.47\textwidth}{!}{

    \begin{tabular}{c | c |c c| c c| c c }
    \hline
     {\multirow{2}{*}{Scales}} & Multi3DRefer & \multicolumn{2}{c|}{ScanQA} & \multicolumn{2}{c|}{SQA3D} & \multicolumn{2}{c}{Scan2Cap} \\
     \cline{2-2} \cline{3-4} \cline{5-6} \cline{7-8}
     & mIoU & C & B-4 & EM & EM-R & C@0.5 & B-4@0.5 \\

    \hline
        
    $r=3$  &40.1	&91.8	&15.1	&51.1	&53.7	&77.0	&35.9	 \\
    
    $r=4$	&44.9&	96.6	&16.6&	54.3&	56.9&	76.6&	35.7 \\
    
    $r=\{2,3\}$  &41.3	&90.6	&15.2	&50.5	&53.3	&75.0	&35.9 \\
    
    $r=\{3,4\}$ &\textbf{46.2}&	94.4&	15.9	&53.9	&56.5&	77.4	&35.8\\
    
    $r=\{2,3,4\}$ & {46.0}	&\textbf{98.6}&	\textbf{17.0}&		{54.4}&	{57.1}&	\textbf{79.0} & 36.7 \\

    $r=\{1,2,3,4\}$ &44.2&	94.7&	16.3	&	\textbf{54.6}&	\textbf{57.3}&	77.1&	35.9  \\

    \hline

    \end{tabular}}

    \caption{Ablation comparing number of scales. Best scores are in \textbf{bold}. Three-scale variant provides a balance between detail preservation and stable reasoning.}
    \label{tab:ablation_num_scales}

\end{table}

\myparagraph{Number of Scales}
We study the impact of the number of scales $R$ used in our multi-scale representation. As shown in \cref{tab:ablation_num_scales}, we compare single-scale, two-scale, three-scale, and four-scale variants. The three-scale configuration yields the strongest overall performance. Single- and two-scale models lack sufficient cross-scale context, limiting their ability to capture both fine geometry and global structure. Increasing to four scales introduces overly fine partitions that contribute noise and lead to mild overfitting, while also incurring higher computational cost. These results confirm that three scales provide an effective balance between detail preservation and stable reasoning. More comparisons are provided in Supplementary Materials.


\begin{table}
\centering
    \setlength{\belowcaptionskip}{-1.7em}
    \centering
    \setlength{\tabcolsep}{1.4pt}
    \resizebox{0.47\textwidth}{!}{

    \begin{tabular}{c | c |c c| c c| c c }
    \hline
     {Number of}
     & Multi3DRefer & \multicolumn{2}{c|}{ScanQA} & \multicolumn{2}{c|}{SQA3D} & \multicolumn{2}{c}{Scan2Cap} \\
     \cline{2-2} \cline{3-4} \cline{5-6} \cline{7-8}
     Queries & mIoU & C & B-4 & EM & EM-R & C@0.5 & B-4@0.5 \\

    \hline
    
    100	&44.3&	90.2 & 14.6&	53.1&55.8&	77.1 & 35.9 \\
    
    200	&45.2&	94.6 & 16.4&	54.2&56.8&	\textbf{79.2} & \textbf{36.7 }\\
    
    400	&\textbf{46.0}&	94.8 & 16.2&	54.0&56.6&	77.4 & 36.3 \\
    
    850	&\textbf{46.0}&	\textbf{98.6} & \textbf{17.0}&	54.4&57.1&	79.0 & \textbf{36.7} \\
    
    1200&	\textbf{46.0}&	96.3 & 16.0&	54.8&57.3&	78.5 & 36.2 \\
    
    1600&	45.9&	95.9 & 16.5&	\textbf{54.9}&\textbf{57.5}&	77.3 & 36.4 \\

    \hline

    \end{tabular}}

    \caption{Ablation comparing number of queries.}
    \label{tab:ablation_num_queries}

\end{table}

\myparagraph{Number of Queries}
\cref{tab:ablation_num_queries} reports the impact of varying the number of learnable queries. Increasing the query count generally improves performance, as a larger query set provides greater expressive capacity and allows MSDec to aggregate richer multi-scale information. Performance saturates around 400–850 queries, indicating that MSDec captures sufficient scene information within this range. Beyond this point, adding more queries yields diminishing returns and can even slightly degrade results, likely due to overfitting to nuanced noise. We adopt 850 queries as it offers the best balance across all benchmarks.

\subsection{Result Visualizations}
\vspace{-0.7em}
We present comparison between our NDTokenizer \includegraphics[height=1.5em]{Figures/robot.png} and 3D-LLaVA \cite{deng20253dllava} \includegraphics[height=1em]{Figures/3dllava.png} in \cref{fig:qualitative} across four tasks.

\myparagraph{Segmentation}
3D-LLaVA often fails to localize all target instances, indicating that its superpoint-pooling–based (downsampling-like) representation loses the fine-grained cues necessary to distinguish nearby or similarly shaped objects. In contrast, NDTokenizer3D accurately segments all referenced instances, benefiting from the preserved local details in our multi-scale NDT representation.

\myparagraph{VQA}
3D-LLaVA tends to rely heavily on language priors rather than the actual 3D scene. In this example, it hallucinates the answer ``mirror", a plausible but incorrect guess given typical sink layouts. NDTokenizer3D instead predicts correctly by grounding its reasoning in the visual tokens.

\myparagraph{SQA}
From the described viewpoint (marked by the red dot), the bathtub is occluded by the wall, leaving the toilet as the only visible candidate. 3D-LLaVA misidentifies the object as its downsampled representation lacks sufficient spatial context. NDTokenizer3D, leveraging information-rich multi-scale features, correctly reasons about visibility and object–environment relationships.

\myparagraph{Captioning}
3D-LLaVA misrecognizes the object and its spatial relation to the environment, again reflecting the limitations of its scene encoding. NDTokenizer3D instead produces correct and spatially coherent descriptions, grounded in the detailed multi-scale representation.

Overall, these qualitative results show that NDTokenizer3D provides more reliable scene representations, capturing fine geometry and broader structure, and achieves more accurate, consistent results across diverse 3D tasks.

\section{Conclusion}
We present NDTokenizer3D, a general-purpose 3D VLM capable of handling a broad range of 3D scene understanding tasks. Our approach introduces a novel three-stage scene tokenization pipeline: (1) constructing a multi-scale Normal Distributions Transform (NDT) representation from high-resolution point clouds, (2) extracting multi-scale 3D features, and (3) aggregating them through the Multi-Scale NDT Decoder (MSDec) to produce holistic scene tokens.
The multi-scale NDT preserves both fine-grained geometric details and global contextual information from raw point clouds without downsampling, forming a compact yet expressive representation of 3D scenes. Beyond feature aggregation, MSDec serves as a unified interface that incorporates human prompts and supports segmentation mask decoding, thereby bridging language-level reasoning with spatial understanding within a single architecture.
Across four 3D vision–language benchmarks, NDTokenizer3D demonstrates impressive performance and broad task coverage, highlighting its effectiveness and versatility in advancing general-purpose 3D vision–language understanding.


\section*{Acknowledgement}
This work was conducted during Yutao’s internship at Microsoft, which also provided computational and research support.

{
    \small
    \bibliographystyle{ieeenat_fullname}
    \bibliography{main}
}

\end{document}